\documentclass{article}

\usepackage{microtype}
\usepackage{graphicx}
\usepackage{subfigure}
\usepackage{booktabs} 
\usepackage{amsmath}
\usepackage{float}

\usepackage{hyperref}

\usepackage{xcolor}
\usepackage{soul}

\usepackage[accepted]{icml2021}

\icmltitlerunning{The Effects of Image Distribution and Task on Adversarial Robustness}

\begin{document}

\twocolumn[
\icmltitle{The Effects of Image Distribution and Task on Adversarial Robustness}



\icmlsetsymbol{equal}{*}

\begin{icmlauthorlist}
\icmlauthor{Owen Kunhardt}{mit}
\icmlauthor{Arturo Deza}{equal,mit}
\icmlauthor{Tomaso Poggio}{equal,mit}
\end{icmlauthorlist}

\icmlaffiliation{mit}{Center for Brains, Minds and Machines (CBMM), Massachusetts Institute of Technology, Cambridge, MA, USA.}

\icmlcorrespondingauthor{Owen Kunhardt}{okunhardt@owenkunhardt.com}

\icmlkeywords{Machine Learning, ICML}

\vskip 0.3in
]



\printAffiliationsAndNotice{\icmlEqualContribution} 

\begin{abstract}
In this paper, we propose an adaptation to the area under the curve (AUC) metric to measure the adversarial robustness of a model over a particular $\epsilon$-interval $[\epsilon_0, \epsilon_1]$ (interval of adversarial perturbation strengths) that facilitates unbiased comparisons across models when they have different initial $\epsilon_0$ performance. This can be used to determine how adversarially robust a model is to different image distributions or task (or some other variable); and/or to measure how robust a model is comparatively to other models. We used this adversarial robustness metric on models of an MNIST, CIFAR-10, and a Fusion dataset (CIFAR-10 + MNIST) where trained models performed either a digit or object recognition task using a LeNet, ResNet50, or a fully connected network (FullyConnectedNet) architecture and found the following: 1) CIFAR-10 models are inherently less adversarially robust than MNIST models; 2) Both the image distribution and task that a model is trained on can affect the adversarial robustness of the resultant model. 3) Pretraining with a different image distribution and task sometimes carries over the adversarial robustness induced by that image distribution and task in the resultant model;  Collectively, our results imply non-trivial differences of the learned representation space of one perceptual system over another given its exposure to different image statistics or tasks (mainly objects vs digits). Moreover, these results hold even when model systems are equalized to have the same level of performance, or when exposed to approximately matched image statistics of fusion images but with different tasks.
\end{abstract}

\section{Introduction}
Adversarial images are perturbed visual stimuli that can fool a high performing image classifier with carefully chosen noise that is often imperceptible to humans~\citep{szegedy2013intriguing,goodfellow2014explaining}. These images are synthesized using an optimization procedure that maximizes the wrong output class of a model observer, while minimizing any noticeable differences in the image for a reference observer. Understanding why adversarial images exist has been studied extensively in machine learning as a way to explore gaps in generalization~\citep{gilmer2018adversarial,yuan2019adversarial,ilyas2019adversarial}, computer vision with applications to real-world robustness~\citep{dubey2019defense,yin2019fourier,richardson2020bayes}, and recently in vision science to understand similar and divergent visual representations with humans ~\citep{zhou2019humans,feather2019metamers,golan2019controversial,reddy2020biologically,dapello2020simulating}. Thus far there have been gaps in the literature on how natural image distributions and classification task impact the robustness of a model to adversarial images. 

This paper addresses whether training on a specific natural image distribution or task plays a role in the adversarial robustness of a model. \textit{Natural images} are images that are representative of the real world. MNIST, CIFAR-10, and ImageNet are examples of natural image datasets. The goal is to understand what it means for a model to inherently be more adversarially robust to objects vs scenes or objects vs digits, where the latter is addressed in this paper. The thesis of this paper is that both the natural image distribution and task (independently and jointly) play a role in the adversarial robustness of a model trained on them. 

Answering questions about the role the image distribution and task in adversarial robustness could be critical for applications where an adversarial image can be detrimental (e.g. self-driving cars~\citep{lu2017no}, radiology~\citep{hirano2020vulnerability} and military surveillance~\citep{ortiz2018defense,deza2019assessment}). These applications are often models of natural image distributions. Understanding if and how the dataset and task play a role in the adversarial robustness of a model could lead to better adversarial defenses for the aforementioned applications and a better understanding of the existence of adversarial images.

The works most closely related to the thesis of this paper are the following: \citet{ilyas2019adversarial} found that adversarial vulnerability is not necessarily tied to the training scheme, but rather is a property of the dataset. Similarly,~\citet{ding2019on} finds that semantic-preserving shifts on the image distribution could result in drastically different adversarial robustness even for adversarially trained models.

 All work in this paper studying the role of natural image distribution and classification task in the adversarial robustness of a model is empirical. The experiments presented require important performance comparisons. Therefore, we propose an unbiased metric to measure the adversarial robustness of a model for a particular set of images over an interval of perturbation strengths. Using this metric, we compare MNIST and CIFAR-10 models and find that MNIST models are inherently more adversarially robust than CIFAR-10 models. We then create a Fusion dataset ($\alpha$-blend of MNIST and CIFAR-10 images) to determine whether the image distribution or task is causing this difference in adversarial robustness and discover that both play a role. Finally, we examine whether pretraining on one dataset (CIFAR-10 or MNIST), then training on the other results in a more adversarially robust learned representation of the dataset the model is trained on and find that this impacts robustness in unexpected ways.  

\begin{figure*}[t!]
    \centering
    \includegraphics[width=2.0\columnwidth]{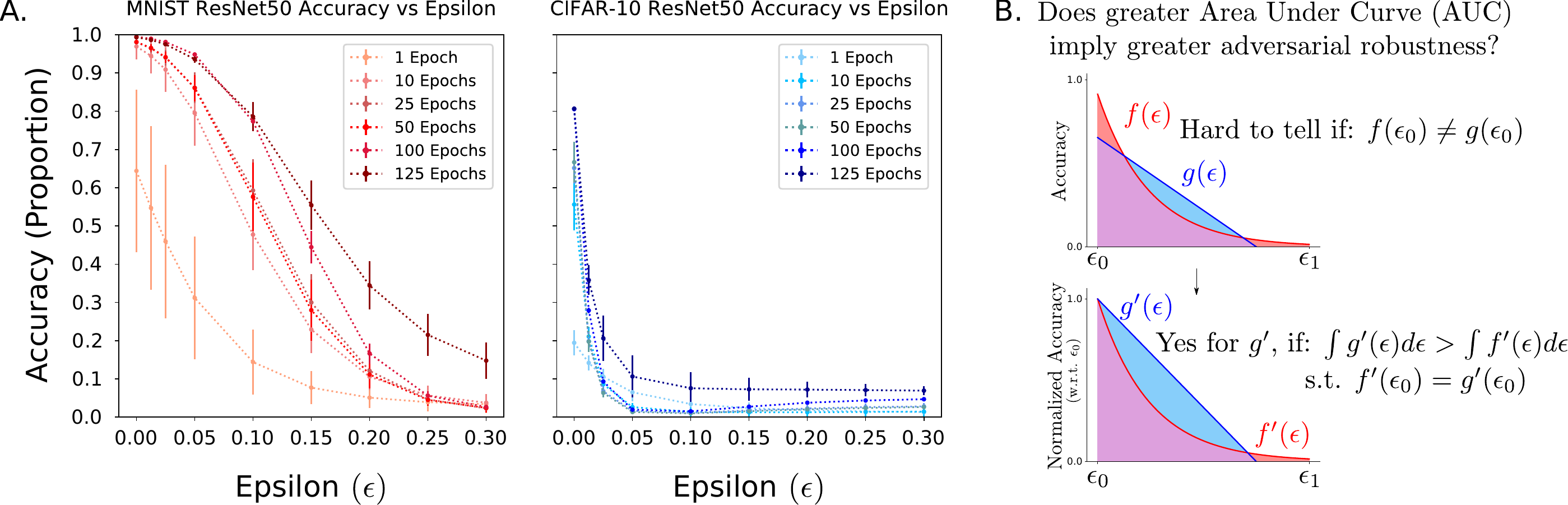}
    \caption{(A) After using the same hyperparameters and training scheme (SGD) for both models, MNIST achieves around $99\%$ accuracy, while CIFAR-10 peaks around $80\%$ with ResNet50 (both without data-augmentation). In cases like these it may be obvious to say that better performing models will be more adversarially robust -- but this is not always the case, in some cases it is the opposite when fixing the image distribution~\citep{zhang2019theoretically}; (B) One solution: example graphs showing the area under the curve of $f(\epsilon)$ and $g(\epsilon$), functions outputting the accuracy of an adversarial attack for a given $\epsilon$ of two models before (top) and after (bottom) accuracy normalization. This shows how at $\epsilon_0$, models go from an unmatched accuracy to a matched upper-bounded score of 1, allowing an unbiased computation of area under the curve.}
    \label{fig:normalized}
\end{figure*}

\section{Proposed Adversarial Robustness Metric}

In order to make comparisons of how robust a model is to adversarial perturbations,  a proper metric for adversarial robustness must be defined. 
We define the adversarial robustness $R$ as a measure of the rate at which accuracy of a model changes as $\epsilon$ (adversarial perturbation strength) increases over a particular $\epsilon$-interval of interest. The faster the accuracy of a model decreases as $\epsilon$ increases, the lower the adversarial robustness is for that model. We propose an adaptation of  area under the curve (AUC) to measure adversarial robustness. A good measure of how much change of accuracy is occurring in an $\epsilon$-interval for a model is the AUC of a function that outputs the accuracy for an adversarial attack given an $\epsilon$ for a model. This AUC provides a total measure of model performance for an $\epsilon$-interval. If the accuracy decreases quickly as $\epsilon$ increases, then the AUC will be smaller. 

Despite how intuitive the previous notion may sound, we immediately run into a problem: Some datasets are more discriminable than others independent of model observers, as shown in Figure~\ref{fig:normalized}(A). This must be taken into account when computing the area under the curve. It could be possible that under unequal initial performances, one model seems more `adversarially robust' over the other by virtue purely of the initial offset in the better performance.

Figure~\ref{fig:normalized}(B) shows that one simple solution to solve the differences in accuracy between two model systems is by normalizing them with respect to their accuracy under non-adversarial ($\epsilon_0=0$) inputs. This yields the following expression: \begin{equation}
\label{eq:Normalization}
    R=\frac{1}{f(\epsilon_0)(\epsilon_1-\epsilon_0)}\int_{\epsilon_0}^{\epsilon_1}f(\epsilon)d\epsilon
\end{equation}
which can be interpreted as the normalized area under the curve of a function $f(\epsilon)$ that outputs the accuracy of a model for an adversarial attack of strength $\epsilon$ over an $\epsilon$-interval (i.e. $[\epsilon_0, \epsilon_1]$). Note that $f(\epsilon_0)>0$ and $\epsilon_1 > \epsilon_0$. Computing $R$ is the same as integrating relative change (shown in supplementary material). Therefore, $R$ is an aggregate measure of relative change in accuracy over an $\epsilon$-interval. The division by $f(\epsilon_0)$ normalizes the function because the function now represents the change in accuracy with respect to no adversarial perturbations (i.e. it is now a relative change). Further, the accuracy at $f(\epsilon_0)$ can be considered an \textit{`oracle'} for the adversarial attacks of the model (i.e. the likely optimal or best performance for that $\epsilon$-interval). The term $\frac{1}{\epsilon_1-\epsilon_0}$ of Eq.~\ref{eq:Normalization} puts the area under the curve of the normalized accuracy between $(0,1]$. This is so that it is easier to interpret and so that the metric is normalized for different $\epsilon$-intervals (i.e. the maximum value is not $\epsilon_1 - \epsilon_0$, but instead is 1). Note that the metric is valid independent of the adversarial attack method.

If for a particular model, $R=1$, this implies that $f(\epsilon)$ is constant over $[\epsilon_0, \epsilon_1]$. If for a model, $R\approx 0$, that means that for all $\epsilon$ in the interval, the model classifies nearly all the perturbed images of a given set incorrectly. $R$ can be arbitrarily close to 0.

To guarantee that $R\leq 1$, the following constraint must be satisfied:  
\begin{equation}
    \int_{\epsilon_0}^{\epsilon_1}f(\epsilon)d\epsilon \leq f(\epsilon_0)(\epsilon_1 - \epsilon_0)
\end{equation}
This is a reasonable constraint to make. An interpretation of $f(\epsilon_0)(\epsilon_1 - \epsilon_0)$ is a possible AUC for $f(\epsilon)$. This AUC occurs when $f(\epsilon) = f(\epsilon_0$) for all $\epsilon\in[\epsilon_0, \epsilon_1]$. In other words, as $\epsilon$ increases, the classification performance of the adversarial images does not change. An AUC greater than $f(\epsilon_0)(\epsilon_1 - \epsilon_0)$ would imply that the accuracy increases above the starting accuracy (i.e. $f(\epsilon_0)$). This behavior would contradict what it means to perform an adversarial attack.

To measure the impact $C$, that adversarial attacks have on a model between two specific $\epsilon$ points instead of an interval, the following can be used: \begin{equation}
\label{eq:relative}
    C =\frac{f(\epsilon)-f(\epsilon_0)}{f(\epsilon_0)}
\end{equation}

where $C$ is the relative change between the performance of a model for two different $\epsilon$'s of adversarial attacks. Normalizing to compute $R$ by taking the relative change in error with respect to a reference or optimal value $f(\epsilon_0)$ (i.e. Eq. \ref{eq:relative}) results in a less biased measure for adversarial robustness than other normalization schemes, such as taking the difference (i.e. $f(\epsilon) - f(\epsilon_0)$). This is because the other schemes are unable to properly account for differences in performance of models on a particular dataset or task. Broadly, we are not interested in how much the performance differs overall, but how much it differs relative from where it started.

There are two methods to find $f(\epsilon)$: 1) to empirically compute multiple values of $\epsilon$ and estimate the area under the curve using integral approximations, such as the trapezoid method; 2) to find the closed form expression of $f(\circ)$ as one would do for psychometric functions~\citep{wichmann2001psychometric} and integrate. In this paper, we do the former (compute multiple values of $\epsilon$ and estimate the integral using the trapezoid method), although this method is extendable to the latter.

Picking $\epsilon_0$ and $\epsilon_1$ is an experimental choice. Choosing $\epsilon_0 = 0$ allows measures the adversarial robustness starting from no perturbations, yet $\epsilon_0 > 0$ can also be used. For too high a choice of $\epsilon_1$, the image can saturate and the performance will likely approach chance. This rebounding effect can be seen in some of the CIFAR-10 curves in our experiments.

There are certain assumptions for this normalization scheme to hold. For example, in both of our experiments MNIST and CIFAR-10 are equalized to have 10 classes and we assume an independent and identically distributed testing distribution such that chance performance for any model observers is the same at $10\%$. One could see how the normalization scheme would give a misleading result if one dataset has 2 i.i.d classes that yield 50\% chance and another dataset yields 10 i.i.d classes that yield 10\% chance. In this case, proportions correct are not comparable and a more principled way of equalizing performance -- likely using $d'$ (a generalized form of Proportion Correct used in Signal Detection Theory) would be required~\citep{green1966signal}.

Overall, this robustness metric can be used to get a sense of whether a model is adversarially robust over a particular $\epsilon$-interval or to measure how adversarially robust a model is comparatively to other models over that interval for a particular set of inputs. Note that this metric is not intended to be used to certify the adversarial robustness of an artificial neural network since it is an approximation of the change of accuracy of a model over an $\epsilon$-interval for, in this paper, a specific set of images.

\begin{figure*}[t!]
    \centering
    \includegraphics[width=2.0\columnwidth]{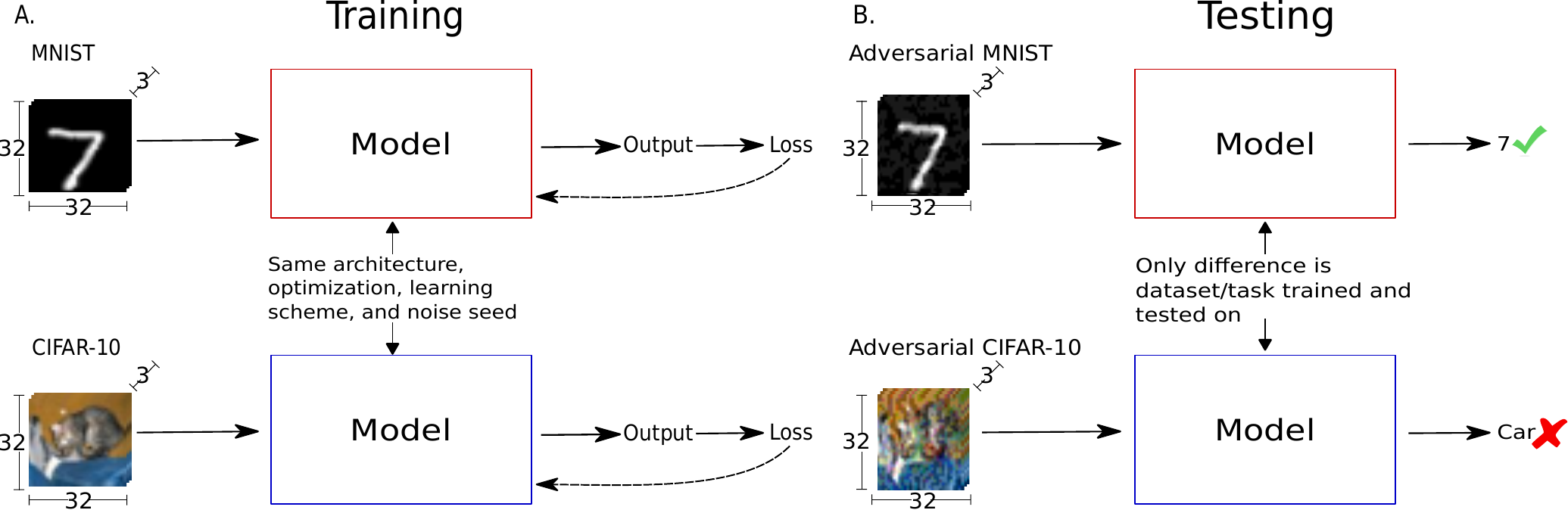}
    \caption{(A) 20 models are trained for each dataset/task (MNIST, CIFAR-10, and later a Fusion Dataset) and network architecture (LeNet, ResNet50, FullyConnectedNet), using a different set of randomly initialized weights (i.e. 60 models per dataset); (B) The models are then tested on adversarial images generated using FGSM \cite{goodfellow2014explaining} of various perturbation strengths. The results from testing result in graphs similar to \ref{fig:normalized}(A). Using these results, the adversarial robustness is computed using Eq. \ref{eq:Normalization}. The average adversarial robustness across the set of two models is compared to determine which model is more adversarially robust and analyze these results.}
    \label{fig:setup}
\end{figure*}

\section{Experimental Design}
\label{sec:Methods}

Figure $\ref{fig:setup}$ visualizes the general experimental design, where models are trained on either MNIST or CIFAR-10 images, and later Fusion images. The architecture, optimization and learning scheme, and initial random weights between each MNIST and CIFAR-10 model is the same, allowing us to draw comparisons between the adversarial robustness of the models after attacking the trained models. 

\subsection{Architectures}
All experiments used 3 networks: LeNet \cite{lenet}, ResNet50 \cite{resnet}, and a fully connected network (FullyConnectedNet) where we explored adversarial robustness over 20 paired network runs and their learning dynamics. FullyConnectedNet has 1 hidden layer with 7500 hidden units. This number of hidden units was chosen so the number of parameters for the FullyConnectedNet has the same order of magnitude as the number of parameters for ResNet50. FullyConnectedNet only have 1 hidden layer so that the network is not biased to approximate a hierarchical function as a convolutional neural network (See~\cite{mhaskar2016deep,poggio2017and} and recently \cite{neyshabur2020towards,deza2020hierarchically}).

\subsection{Datasets}
The datasets used were MNIST, CIFAR-10, and a Fusion Dataset. To use the exact same architectures with the datasets, MNIST was upscaled to $32\times32$ and converted to 3 channels to match the dimensions of CIFAR-10 (i.e. $32\times 32 \times 3$). MNIST was changed instead of CIFAR-10 because given the low image complexity of MNIST images  -- mainly their low spatial frequency structure, that lends itself to upscaling -- and thus it would be less likely to change the accuracy of models trained on that dataset. Preliminary results showed that there is a difference (insignificant in comparison to the other differences in results in this paper) in the adversarial robustness of models trained on the scaled and 3 color channel version and the regular version, with the scaled and 3 color channel version being less robust. Whether the changes to MNIST entirely caused the difference was not determined due to the differences between architectures that were used for each version. No other changes to the datasets were made (such as color normalization, which is typically used for CIFAR-10) in order to preserve the natural image distribution.

\begin{figure}[!t]
    \centering
    \includegraphics[width=1.0\columnwidth]{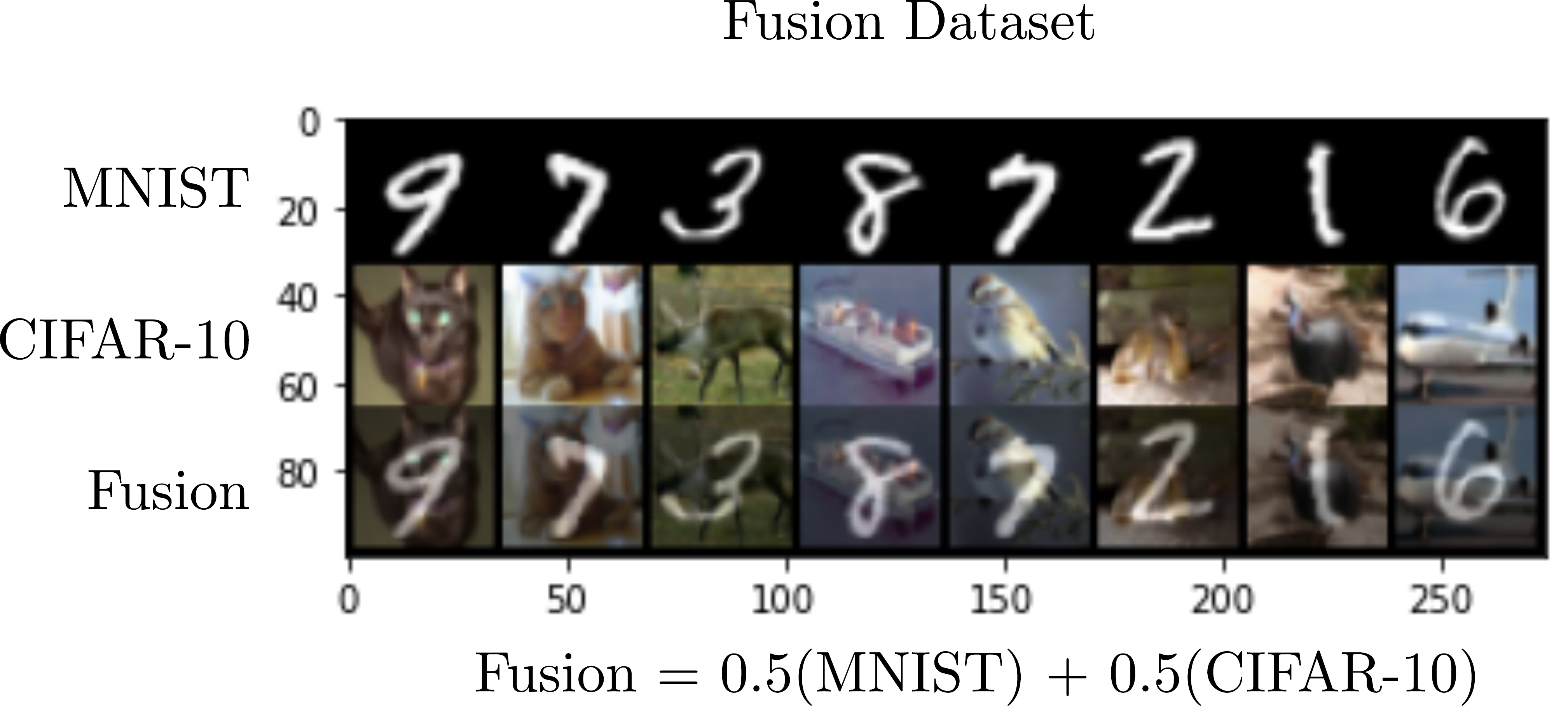}
    \caption{The Fusion dataset was created to tease apart the cause and effects of the inherit adversarial robustness of models trained on natural image distributions. Here, we show sample images from the Fusion dataset consisting of alpha-blended MNIST + CIFAR-10 stimuli.}
    \label{fig:fusion}
\end{figure}

The Fusion dataset that is used in the experiments is not a natural image distribution. It was created with the purpose of better understanding the inherit adversarial robustness proprieties of natural image distribution models. Each fusion image in the dataset is generated with the following $\alpha$-blending procedure: 
\begin{equation}
\label{eq:fusion}
    F = 0.5M + 0.5C,
\end{equation} where $F$ is a new fusion image, $M$ is an MNIST image modified to 32x32x3 (by upscaling and increasing number of color channels), and $C$ is a CIFAR-10 image. Example fusion images can be found in Figure~\ref{fig:fusion}. This dataset is similar to \textit{Texture~shiftMNIST} from~\citet{jacobsen2018excessive}.

The Fusion dataset was created online during training or testing during each mini-batch by formula~\ref{eq:fusion}. The fusion image training set was constructed using the MNIST and CIFAR-10 training set and the fusion image test set was constructed using the MNIST and CIFAR-10 test set. During training, the MNIST and CIFAR-10 datasets are shuffled at the start of every epoch. Therefore, it is likely that no fusion images are shown to the model more than once. This was done to ensure that the model cannot learn any correlation between any CIFAR-10 object and any MNIST digit, as well as, improve generalization of the model. Additionally, it is important to note that no two models were trained on the exact same set of fusion images, but were evaluated on the same test images. Since we train 20 random models, it should average out any possible noise to a certain degree, but strictly speaking the images were different but the statistics were approximately matched.

\subsection{Hyperparameters, Optimization Scheme, and Initialization}
It is important to note that all hyperparameters are held constant, including the $\epsilon$-interval. The only difference between the models using a certain architecture is the dataset/task they are trained and tested on (just the task in the case of the Fusion dataset). In the experiments presented, the independent variables are the dataset and task, while the dependent variable being measured is the adversarial robustness of the model. Since all other variables are held fixed, if the adversarial robustness of the models trained on the different datasets/tasks are different, then this change is due to the dataset/task itself (i.e. the image distribution and classification task). If the $\epsilon$-interval used to attack the two models is different we could not directly conclude that any differences are due to the image distribution and task because the difference could also be due to the differences in the strengths of the adversarial attacks on each model. Experiments using the Fusion dataset are presented in this paper to investigate which of the independent variables (i.e. whether image distribution or task) is playing a role in the differences in adversarial robustness.

The loss function used for all models was cross-entropy loss and the optimizer used was stochastic gradient descent (SGD) with weight decay $5\times10^{-4}$, momentum $0.9$, and with an initial learning rate 0.01 for the FullyConnectedNet and LeNet models and an initial learning rate 0.1 for the ResNet50 models. The learning rate was divided by 10 at 50\% of the training. The FullyConnectedNet and LeNet models were trained to 300 epochs and the ResNet50 models were trained to 125 epochs. ResNet50 models required less epochs during training because those models reached high levels of performance sooner than the other architectures.  A batch size of 125 was used. The batch size was 125 since this is the closest number to a more typical batch size of 128 that divides both the number of CIFAR-10 images and MNIST images. This was needed to ensure that the batches align properly when creating the fusion images. These hyperparameters and optimization scheme were chosen since they resulted in the best performance of those tested in preliminary experiments. 

For all experiments, each model was trained 20 times with matched initial random weights across different datasets. For example in the case of LeNet, 20 different LeNet models all with different initial random weights:~$\{w_1,w_2,...,w_{20}\}$ were used to train for CIFAR-10 in our first experiment, and these same initial random weights were used to train for MNIST. This removed the variance induced by a particular initialization (\textit{e.g.} a lucky/unlucky noise seed) that could bias the comparisons by arriving to a better solution via SGD. This procedure was possible because our MNIST dataset was resized to a 3-channeled version with a new size of $32\times32\times3$ instead of $28\times28\times1$ (original MNIST).

\begin{figure*}[t]
    \centering
   \includegraphics[width=2.0\columnwidth]{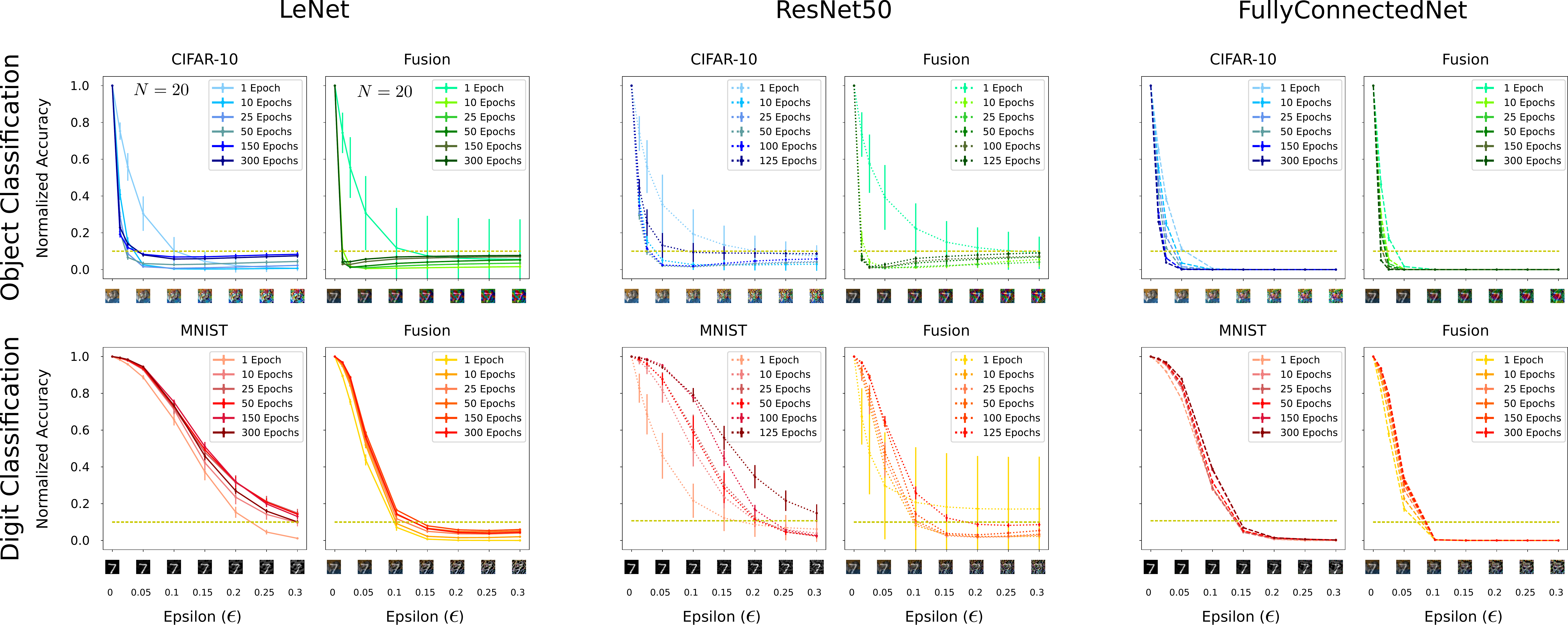}
    \caption{MNIST-trained networks (bottom left) across all architectures show greater adversarial robustness after accuracy normalization than CIFAR-10 trained networks (top left for each architecture). Notice too that ResNet50 appears to be the more adversarially robust network across network architectures (LeNet and FullyConnectedNet) independent of learning dynamics. Graphs of the normalized accuracy of the Fusion dataset on the object recognition task (top right) and digit recognition task (bottom right) for LeNet, ResNet50, and FullyConnectedNet. Generally, models trained on the digit task were more adversarially robust than those trained on the object task, showing the role that task plays in the adversarial robustness of a model. Additionally, these models were generally less adversarially robust than their MNIST and CIFAR-10 model counterparts. In combination, these results imply that both task and image distribution play distinct roles in the adversarial robustness of a model. The gold lines represent chance performance in the graphs.}
    \label{fig:MNISTvsCIFAR_Original_and_Fusion}
\end{figure*}

\subsection{Adversarial Attacks}

The method used for generating adversarial images in the experiments presented in this paper is the Fast Gradient Sign Method (FGSM) presented in \citet{goodfellow2014explaining}. The focus of the attacks was to create images that cause the model to misclassify in general, rather than misclassifying an image to a particular class. FGSM was chosen over other optimization based attacks such as Projected Gradient Descent (PGD) \cite{madry2019deep} based on preliminary results as FGSM was sufficient to successfully adversarially attack the model. FGSM also has a lower computational cost than PGD allowing us to run more experiments and train more models. Adversarial training or other data-augmentations schemes that may bias the outcome were not performed. Importantly, given that an adversarial defense mechanism is not being proposed or used, strong adversarial attack methods, such as PGD, are not necessary in this first work -- contrary, but justified to the advice from \citet{carlini2019evaluating}.

The $\epsilon$-interval used in the experiments is $[0,0.3]$ (i.e. $\epsilon_0 = 0, \epsilon_1 = 0.3$). The upper bound of $0.3$ was chosen because adversarial images at that magnitude are difficult for many undefended classifiers to classify correctly. The trained models were adversarially attacked with $\epsilon\in\{0, 0.0125, 0.025, 0.05, 0.1, 0.15, 0.2, 0.25, 0.3\}$ to approximate $f(\epsilon)$. For the models using LeNet and FullyConnectedNet architectures, they were adversarially attacked at 1, 10, 25, 50, 150, and 300 epochs. Models using the ResNet50 architecture were adversarially attacked at 1, 10, 25, 50, 100, and 125 epochs. Different epochs were adverarially attacked to determine whether the results differed at different stages of learning.

\section{Experimental Results}

The following experiments provide a glimpse into the role of classification task and image distribution in the adversarial robustness of models.

All differences in robustness that are mentioned are statistically significant using a Welch's t-test with significance level $\alpha=0.05$. This test was used because the models are unpaired and do not have equal variance since the models are trained on different datasets.

\subsection{Comparing MNIST vs CIFAR-10 Adversarial Robustness}
This experiment investigates whether MNIST models are inherently more adversarially robust than CIFAR-10 models. This was investigated by comparing the adversarial robustness of CIFAR-10 models and the MNIST models for the three architectures. Figure~\ref{fig:MNISTvsCIFAR_Original_and_Fusion} (top left for each architecture) shows normalized accuracy graphs for the CIFAR-10 trained models and  Figure~\ref{fig:MNISTvsCIFAR_Original_and_Fusion} (bottom left) shows graphs of normalized accuracy for MNIST trained models. Both LeNet and FullyConnectedNet, the MNIST models were more adversarially robust than CIFAR-10 models, for each epoch we examined. The same pattern of results held for ResNet50 models except for the first epoch where there was no difference between the MNIST and CIFAR-10 models.

\underline{Result 1:} For the three network architectures tested (that all vary in approximation power and architectural constraints), MNIST trained models are inherently more adversarially robust than CIFAR-10 models. This implies that the task and/or image distribution play a role in the adversarial robustness of the model.

\subsection{Comparing Object vs Digit Classification in the Fusion (MNIST + CIFAR-10)  dataset}

The previous results suggested that after taking into account different measures of accuracy normalization, MNIST (both dataset and digit recognition task) models are intrinsically more adversarially robust than CIFAR-10 models. This implies that it is harder to fool an MNIST model, than a CIFAR-10 model, likely, in part, due to the fact that number digits are highly selective to shape, and show less perceptual variance than objects.

Naturally, the next question that arises is if the task itself is somehow making each perceptual system less adversarially robust. To test this hypothesis the Fusion dataset was used. Models were trained to perform either digit recognition or object recognition on these fusion images -- thus we have approximately fixed the image distribution but varied the approximation task~\citep{deza2020hierarchically}. They are approximately matched because no model is trained on the exact same images, the image distribution is approximately the same on average given the random sampling procedure. The goal with this new hybrid dataset is to re-run the same set of previous experiments and test adversarial robustness for both the digit recognition task and the object recognition task and probe the role
of the type of classification task when fixing the dataset to test how adversarial robustness varies when all other variables remain constant.

Observation: When examining the first epoch for the fusion trained models, the standard deviation of the curves in \ref{fig:fusion}(B) are generally high. This is likely due to design choice of avoiding to show the same fusion image twice. This does not occur in later stages of training.

\underline{Result 2a:} Task plays a critical role in the adversarial robustness of a model. Figure ~\ref{fig:MNISTvsCIFAR_Original_and_Fusion} contains the normalized curves of the results for the digit and object recognition tasks on the fusion dataset for each of the architectures. The models were evaluated on fusion images constructed from the MNIST and CIFAR-10 test sets. The FullyConnectedNet (all epochs), ResNet50 and LeNet fusion image models were more adversarially robust on the digit recognition task than the object recognition task for all epochs examined excluding the first epoch. This suggests that even if the image distribution is approximately equalized at training, the representation learned varies given the task, and impacts adversarial robustness differently.

\underline{Result 2b:} Image distribution also plays a role in the adversarial robustness of a model. Comparing the three architectures trained on the Fusion Dataset vs their regular image-distribution trained models show that increasing the image complexity (by adding a conflicting image with the hope of increasing invariance) in fact decreases adversarial robustness when compared to regularly trained networks. Comparing fusion image models trained on the digit task and MNIST models: for the FullyConnectedNet and LeNet architecture, the MNIST models were more robust. The same holds for the ResNet50 MNIST models except at the first epoch, where there was no difference. CIFAR-10 models using the FullyConnectedNet architecture were more adversarially robust than the fusion image models trained on the object recognition task for all epochs tested. The same was true for the LeNet and ResNet50 architectures except there were no differences between CIFAR-10 models and fusion images with object task in adversarial robustness for 1 and 50 epochs. 

\begin{figure*}[t]
    \centering
   \includegraphics[width=2.0\columnwidth]{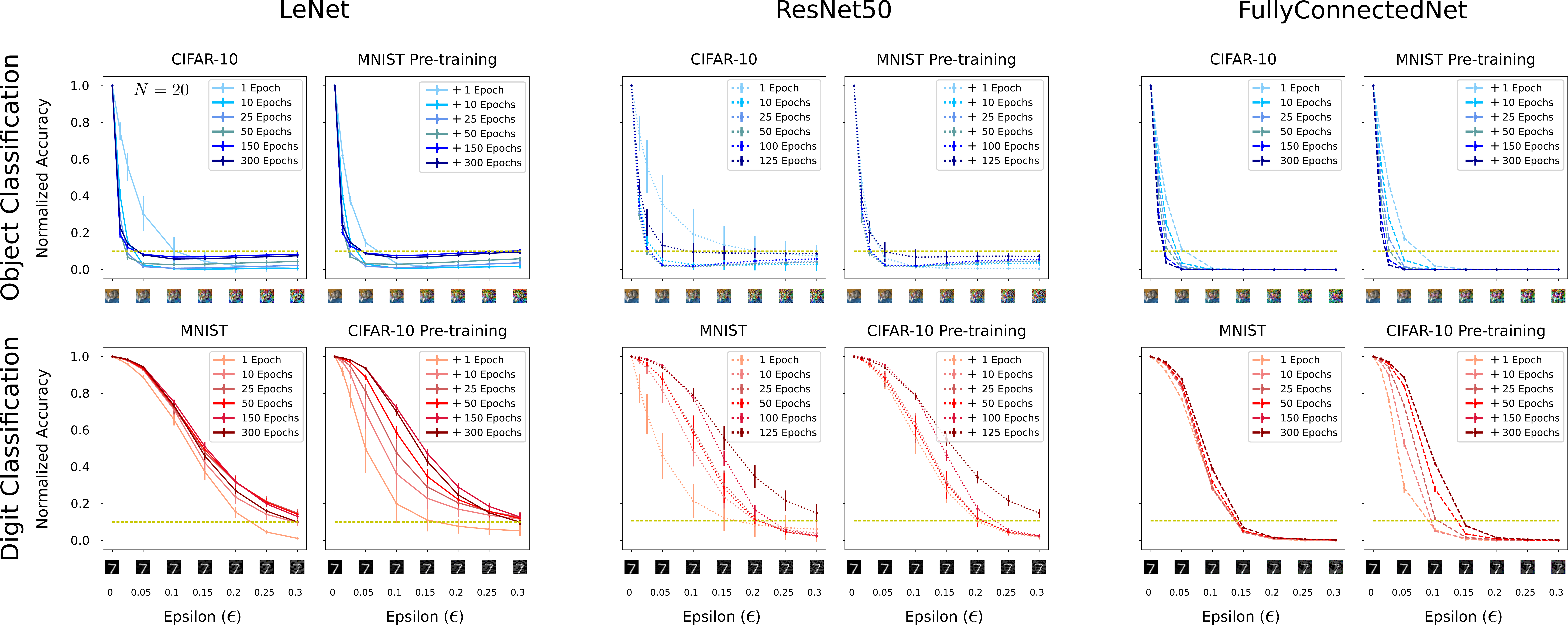}
    \caption{\textit{Visual Hysteresis}: 
    FullyConnectedNet and LeNet networks seems to carry over the learned representation and adversarial vulnerability from the pretrained system. However, only LeNet experiences a clear visual hysteresis where pretraining on CIFAR-10 for MNIST is worse (less adversarially robust) than only training on MNIST, yet pretraining on MNIST for CIFAR-10 is better (more adversarially robust) than only training on CIFAR-10 (See supplementary material). The gold lines represent chance performance in the graphs.}
    \label{fig:MNISTvsCIFAR_Original_and_Pretraining}
\end{figure*}

\subsection{Impact of Pretraining on Out-Of-Distribution (o.o.d) image datasets}

This experiment investigates whether pretraining on one dataset (CIFAR-10 or MNIST), then training on the other results in a more adversarially robust learned representation of the dataset the model is trained on. 

The pretraining procedure was done by using the existing fully trained CIFAR-10 or MNIST FullyConnectedNet, LeNet, and ResNet50 models as bases and then training/fine-tuning them using the same training scheme but with MNIST or CIFAR-10 respectively. These models were then tested using the test sets of the datasets the models were fine-tuned on.

For the FullyConnectedNet, the MNIST models were more adversarially robust than the MNIST pretrained on CIFAR-10 model during early stages of learning, but the pretrained models were more robust when examined at 150 and 300 epochs of fine-tuning. The MNIST LeNet models were more adversarially robust for all stages of learning than the pretrained model. The pretrained ResNet50 models had no differences in robustness compared to the MNIST ResNet50 models, except for the first epoch where the pretrained models were more robust. This result is unexpected as this does not occur for the other architectures. These results would seem to suggest that architecture plays a role in the adversarial robustness of the learned representation contingent on the given dataset/task and potentially compositional nature. 

Pretraining on CIFAR-10 and then training on MNIST generally does not lead to more adversarially robust models. Next we investigate whether pretraining on MNIST and then training on CIFAR-10 has this same effect. We find that this is not always the case. Pretraining on MNIST then training on CIFAR-10 led to marginal improvements in adversarial robustness for LeNet, except for the first epoch (Figure~\ref{fig:MNISTvsCIFAR_Original_and_Pretraining}). For ResNet50, pretraining resulted in less adversarially robust models at the start and end of training (1 and 125 epochs), otherwise there was no difference compared to not pretraining. The FullyConnectedNet pretrained models were more adversarially robust in earlier stages of learning, but were less robust in later stages. Tables of the robustness metrics for the CIFAR-10 models pretrained on MNIST (as well as for other experiments) can be found in the supplementary material. These findings requires further investigation.   

For the ResNet50, LeNet, and FullyConnectedNet architectures, the models pretrained on CIFAR-10 then trained on MNIST were statistically significantly more adversarially robust than models pretrained on MNIST then trained on CIFAR-10 for all epochs examined.

\underline{Result 3}: Pretraining on CIFAR-10 followed by training on MNIST does not generally produce a more adversarially robust model than training on MNIST alone, with any of the tested architectures. This is counter intuitive given that humans typically base their learned representations on objects rather than figures~\citep{janini2019shape}. On the other hand, pretraining on MNIST, then training on CIFAR-10 only aided LeNet; for FullyConnectedNet it helped in earlier stages of learning, while decreased robustness later. Generally, however, ResNet50 models were not affected in terms of carried-over robustness at any intermediate stages of learning. Investigating the origins of this visual hysteresis (an asymmetry in learned representation visible through robustness given the pretraining scheme)~\citep{sadr2004object} and how it may relate to shape/texture bias~\citep{geirhos2018imagenet,hermann2019exploring}, spatial frequency sensitivity~\citep{dapello2020simulating,deza2020emergent}, or common perturbations~\citep{hendrycks2018benchmarking} is a subject of on-going work.

\section{Discussion}
This work verified that both the image distribution and task (independently or jointly) can impact the adversarial robustness of a model under FGSM. The next step is to investigate why, and what specific factors of the image statistics and task play a role. It is likely that MNIST trained networks are intrinsically more adversarially robust than CIFAR-10 trained networks in part due to the lower-dimensional subspace in which they live in given their image structure~\citep{henaff2014local} compared to CIFAR-10 (\textit{i.e.} MNIST has less non-zero singular values than CIFAR-10 allowing for greater compression for a fixed number of principal components). Additionally, in future work we want to know whether these observations hold with other optimization based attacks and gradient-free attacks, such as PGD \cite{madry2019deep} and NES \cite{ilyas2018blackbox} respectively. Given that FGSM is not considered a strong attack, would a stronger attack exacerbate these results? Based on the noticeable differences in adversarial robustness between the models testing only using FGSM, this is a promising direction. 

Indeed, this paper has only scratched the surface of the role of natural image distribution and task in the adversarial robustness of a model by comparing two well known candidate datasets over their learning dynamics: MNIST and CIFAR-10. Continuing this line of work onto exploring the role of the image distribution on adversarial robustness for other natural image distributions such as textures or scenes is another promising next step. Finally, future experiments should continue to investigate the effect of the learning objective on the learned representation induced from the image distribution. We have already seen how the task affects the adversarial robustness of a model even when image statistics are approximately matched under a supervised training paradigm. With the advent of self-supervised~\citep{konkle2020instance,geirhos2020on,purushwalkam2020demystifying} and unsupervised~\citep{zhuang2020unsupervised} objectives that may be predictive of human visual coding, it may be relevant to investigate the changes in adversarial robustness for the current (objects, digits) and new (texture, scenes) image distributions with the proposed adversarial robustness metric for these new learning objectives. 

\section{Acknowledgements}
The authors thank Dr. Christian Bueno and Dr. Susan Epstein for their helpful feedback on this paper. This work was supported in part by the Center for Brains, Minds and Machines (CBMM), funded by NSF STC award  CCF – 1231216.

\bibliography{robustness_paper}
\bibliographystyle{icml2021}

\cleardoublepage
\newpage
\appendix
\section{Appendix}
\subsection{Metric}
\label{sec:metric}
The relative change between the accuracy of a model for two different $\epsilon$'s is Eq. \ref{eq:relative}. We will now show that the robustness $R$ (Eq. \ref{eq:Normalization}) is indeed a measure relative change.   

$
    \frac{1}{\epsilon_1 - \epsilon_0}\int_{\epsilon_0}^{\epsilon_1}\frac{f(\epsilon)-f(\epsilon_0)}{f(\epsilon_0)}d\epsilon = \frac{1}{\epsilon_1 - \epsilon_0}\int_{\epsilon_0}^{\epsilon_1}\left(\frac{f(\epsilon)}{f(\epsilon_0)}-\frac{f(\epsilon_0)}{f(\epsilon_0)}\right)d\epsilon
    \\\\
    = \frac{1}{\epsilon_1 - \epsilon_0}\left(\int_{\epsilon_0}^{\epsilon_1}\frac{f(\epsilon)}{f(\epsilon_0)}d\epsilon-\int_{\epsilon_0}^{\epsilon_1}\frac{f(\epsilon_0)}{f(\epsilon_0)}d\epsilon\right)
    \\\\
    = \frac{1}{\epsilon_1 - \epsilon_0}\left(\frac{1}{f(\epsilon_0)}\int_{\epsilon_0}^{\epsilon_1}f(\epsilon)d\epsilon-\int_{\epsilon_0}^{\epsilon_1}1d\epsilon\right)
    \\\\
    = \frac{1}{\epsilon_1 - \epsilon_0}\left(\frac{1}{f(\epsilon_0)}\int_{\epsilon_0}^{\epsilon_1}f(\epsilon)d\epsilon-(\epsilon_1 - \epsilon_0) \right)
    \\\\
    = \frac{1}{f(\epsilon_0)(\epsilon_1 - \epsilon_0)}\int_{\epsilon_0}^{\epsilon_1}f(\epsilon)d\epsilon- \frac{\epsilon_1 - \epsilon_0}{\epsilon_1 - \epsilon_0}
    \\\\
    = \frac{1}{f(\epsilon_0)(\epsilon_1 - \epsilon_0)}\int_{\epsilon_0}^{\epsilon_1}f(\epsilon)d\epsilon- 1$

Notice that the output of this is in $(-1,0]$, where values close to $-1$ imply low adversarial robustness and $0$ implies a model did not change in classification accuracy at all as $\epsilon$ increases. Therefore, ignoring the shift of -1 so that the measure is easier to interpret and more intuitive, we get $R$.

\cleardoublepage

\subsection{Unnormalized Adversarial Robustness Curves (Raw-Data)}

 \begin{figure}[!t]
    \centering
    \twocolumn[{\includegraphics[width=2.0\columnwidth]{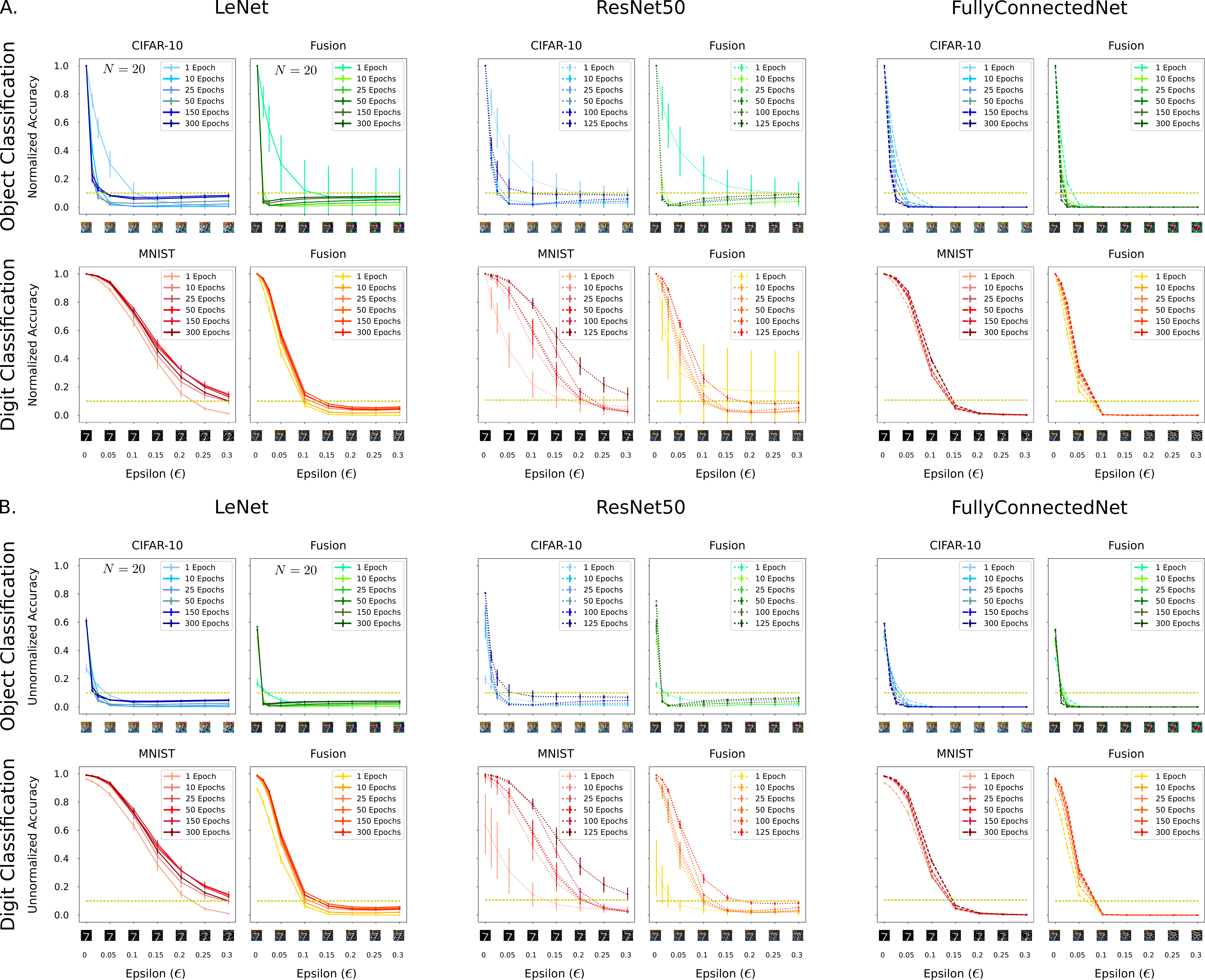}\caption{(A): Redrawn graphs from Figure~\ref{fig:fusion} as a reference; (B): The un-normalized adversarial robustness trade-off curves for each network (LeNet, ResNet50, FullyConnectedNet) and dataset (MNIST and CIFAR-10) for the FGSM-based Attack~\citep{goodfellow2014explaining}.}
    \label{fig:Raw_Fusion}}]
\end{figure}

 Figure~\ref{fig:Raw_Fusion}A are redrawn graphs from Figure~\ref{fig:fusion} as a reference.  Figure~\ref{fig:Raw_Fusion}B are the unnormalized versions of those graphs. In other words, the points in the normalized graphs are not a proportion out of $f(\epsilon_0)$ and are the raw accuracies for the models at each $
 \epsilon$.
 
 \underline{Case study:} In Figure~\ref{fig:Raw_Fusion}B, notice that for LeNet, ResNet50, and FullyConnectedNet (with the exception of epoch 1 for the ResNet50 models), the models for the Fusion dataset with the digit recognition task and MNIST have approximately the same average initial performance for each epoch examined. Observe that although starting at similar points, Fusion digit recognition models accuracies decreases faster as $\epsilon$ increases than the MNIST models. We find that indeed MNIST models are more adversarially robust than Fusion digit recognition models. Here, it is clear to see without the normalization.
 
\begin{figure}[!t]
    \centering
    \twocolumn[{\includegraphics[width=2.0\columnwidth]{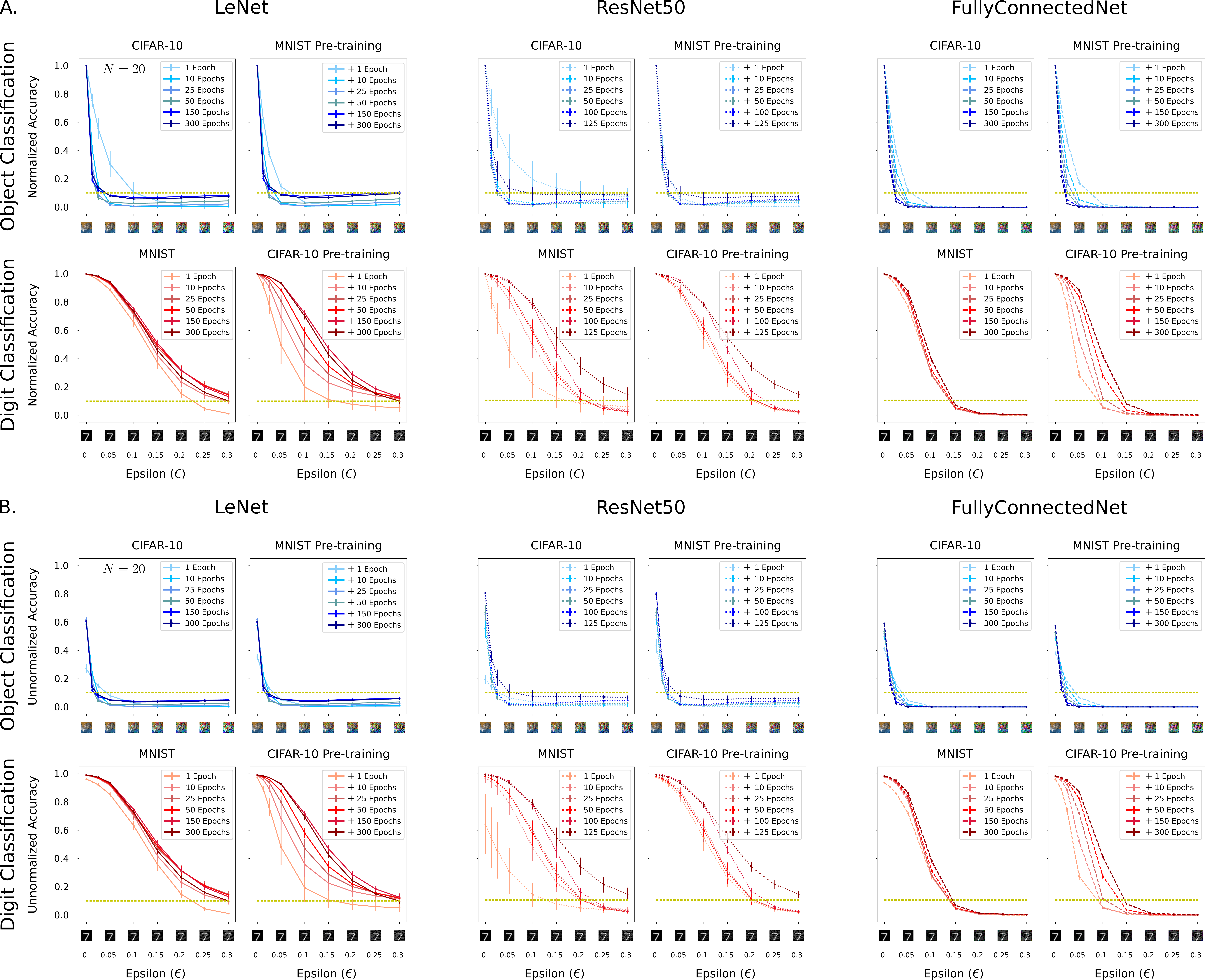}\caption{(A): Redrawn graphs from Figure~\ref{fig:MNISTvsCIFAR_Original_and_Pretraining}; (B): The un-normalized adversarial robustness trade-off curves for each network (LeNet, ResNet50, FullyConnectedNet) and Fusion dataset for the FGSM-based Attack~\citep{goodfellow2014explaining}.}
    \label{fig:PreTraining_Raw}}]
\end{figure}

 Figure~\ref{fig:PreTraining_Raw}A are redrawn graphs from Figure~\ref{fig:MNISTvsCIFAR_Original_and_Pretraining} as a reference. Figure~\ref{fig:PreTraining_Raw}B are the unnormalized versions of those graphs.

\cleardoublepage

\cleardoublepage

\subsection{Zoomed in Sample Adversarial Stimuli}

\begin{figure}[!t]
    \centering
    \twocolumn[{\includegraphics[width=1.665\columnwidth]{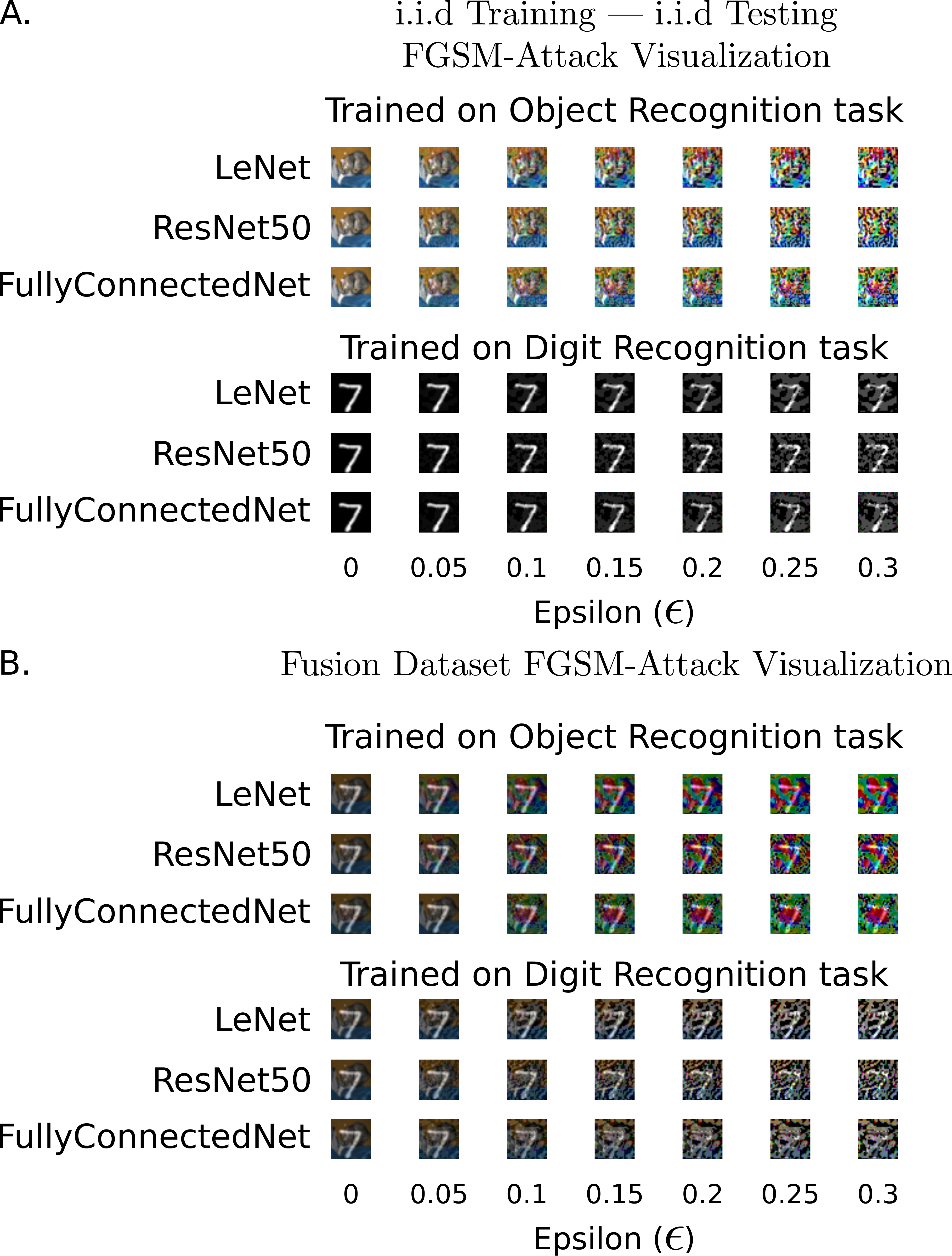}
    \caption{Zoomed in versions of the adversarial patches created after an FGSM-Attack. Shown images are the perturbed stimuli for the networks at the 300th,125th,300th epoch for LeNet, ResNet50 and FullyConnectedNet respectively. A and B show sample stimuli from our first experiment. Interestingly, the differences in the adversarial noise pattern are more salient across architectures for the Fusion Dataset.}
    \label{fig:Zoomed_All_Adversarial}
    }]
\end{figure}

\cleardoublepage

\begin{figure}[!t]
    \centering
    \twocolumn[{\includegraphics[width=1.8\columnwidth]{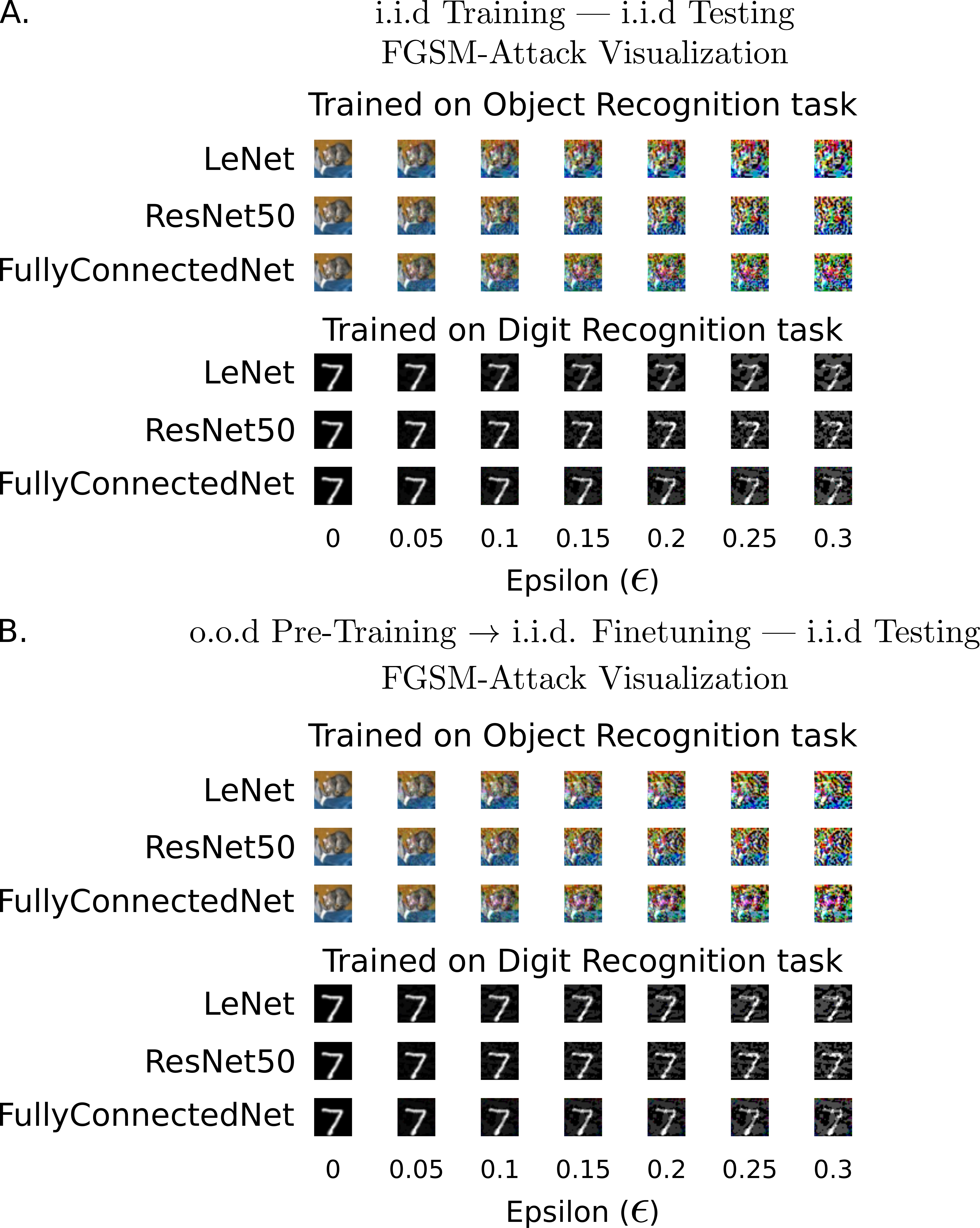}
    \caption{Zoomed in versions of the adversarial patches created after an FGSM-Attack. Shown images are the perturbed stimuli for the networks at the 300th,125th,300th epoch for LeNet, ResNet50 and FullyConnectedNet respectively. A and B show sample stimuli from our second experiment. Interestingly, the differences in the adversarial noise pattern are more salient across architectures for the Fusion Dataset.}
    \label{fig:Zoomed_All_Adversarial}
    }]
\end{figure}

\cleardoublepage

\subsection{Statistical Testing of Results (Extended)}
The tables below contain the robustness $R$ computed using Eq. \ref{eq:Normalization}, averaged over the 20 models for the specified dataset/task and architecture at various stages of learning.

\underline{Table Legend:}
\\
$\star$: Denotes statistically significantly higher adversarial robustness for MNIST vs CIFAR-10
\\
$\bullet$: Denotes statistically significantly higher adversarial robustness for a pretrained MNIST model vs a pretrained CIFAR-10 model
\\
$\diamond$: Denotes statistically significantly higher adversarial robustness for a pretrained model vs non-pretrained model both trained on same dataset
\\
$\dagger$: Denotes statistically significantly higher adversarial robustness for Fusion digit task vs Fusion object task
\\
$\ddagger$: Denotes statistically significantly higher adversarial robustness for MNIST vs Fusion digit task or CIFAR-10 vs Fusion object task
\\
Results from fully trained models are bolded.

\label{sec:Results_Table}
\begin{table}[h]
\begin{center}
\caption{MNIST LeNet adversarial robustness}
\vspace{0.1in}
\label{tab:lenet-mnist}
\begin{tabular}{@{}lll@{}}
\toprule
Epoch: & Mean Robustness: & SD:     \\ \midrule
1      & 0.438451$\star\diamond\ddagger$        & 0.020373 \\ \midrule
10     & 0.499361$\star\diamond\ddagger$       & 0.023713 \\ \midrule
25     & 0.539882$\star\diamond\ddagger$        & 0.025632 \\ \midrule
50     & 0.543602$\star\diamond\ddagger$      & 0.019689 \\ \midrule
150    & 0.549709$\star\diamond\ddagger$        & 0.013361 \\ \midrule
\textbf{300 }   & \textbf{0.520547}$\star\diamond\ddagger$        & \textbf{0.010607} \\ \bottomrule
\end{tabular}
\end{center}
\end{table}

\begin{table}
\begin{center}
\caption{CIFAR-10 LeNet adversarial robustness}
\vspace{0.1in}
\label{tab:lenet-cifar}
\begin{tabular}{@{}lll@{}}
\toprule
Epoch: & Mean Robustness: & SD:     \\ \midrule
1      & 0.155815$\diamond$         & 0.048486 \\ \midrule
10     & 0.053812$\ddagger$        & 0.003744 \\ \midrule
25     & 0.050288$\ddagger$         & 0.00447  \\ \midrule
50     & 0.063126         & 0.009296 \\ \midrule
150    & 0.102256$\ddagger$         & 0.009832 \\ \midrule
\textbf{300}    & \textbf{0.097237}$\ddagger$         & \textbf{0.008319} \\ \bottomrule
\end{tabular}
\end{center}
\end{table}

\begin{table}[h]
\begin{center}
\caption{Pretrained on CIFAR-10 trained on MNIST LeNet adversarial robustness}
\vspace{0.1in}
\label{tab:lenet-pre-cifar}
\begin{tabular}{@{}lll@{}}
\toprule
Epoch: & Mean Robustness: & SD:      \\ \midrule
1      & 0.249363$\bullet$         & 0.05819  \\ \midrule
10     & 0.364303$\bullet$         & 0.049268 \\ \midrule
25     & 0.420838$\bullet$         & 0.039604 \\ \midrule
50     & 0.462459$\bullet$         & 0.022959 \\ \midrule
150    & 0.529665$\bullet$         & 0.013517 \\ \midrule
\textbf{300}    & \textbf{0.503499}$\bullet$         & \textbf{0.013186} \\ \bottomrule
\end{tabular}
\end{center}
\end{table}
\begin{table}[h]
\begin{center}
\caption{Pretrained on MNIST trained on CIFAR-10 LeNet adversarial robustness}
\vspace{0.1in}
\label{tab:lenet-pre-mnist}
\begin{tabular}{@{}lll@{}}
\toprule
Epoch: & Mean Robustness: & SD:      \\ \midrule
1      & 0.102636         & 0.008461 \\ \midrule
10     & 0.059423$\diamond$         & 0.005675 \\ \midrule
25     & 0.057364$\diamond$         & 0.008041 \\ \midrule
50     & 0.069428$\diamond$         & 0.006776 \\ \midrule
150    & 0.113739$\diamond$         & 0.007872 \\ \midrule
\textbf{300}    & \textbf{0.109077}$\diamond$         & \textbf{0.008987} \\ \bottomrule
\end{tabular}
\end{center}
\end{table}

\begin{table}[h]
\begin{center}
\caption{Fusion digit task LeNet adversarial robustness}
\vspace{0.1in}
\label{tab:lenet-digit}
\begin{tabular}{@{}lll@{}}
\toprule
Epoch: & Mean Robustness: & STD:     \\ \midrule
1      & 0.173351        & 0.008095 \\ \midrule
10     & 0.204077$\dagger$         & 0.006719 \\ \midrule
25     & 0.225057$\dagger$         & 0.008112 \\ \midrule
50     & 0.238905$\dagger$         & 0.0096   \\ \midrule
150    & 0.254536$\dagger$         & 0.005374 \\ \midrule
\textbf{300}    & \textbf{0.239281}$\dagger$         & \textbf{0.004036} \\ \bottomrule
\end{tabular}
\end{center}
\end{table}
\begin{table}[h]
\begin{center}
\caption{Fusion object task LeNet adversarial robustness}
\vspace{0.1in}
\label{tab:lenet-object}
\begin{tabular}{@{}lll@{}}
\toprule
Epoch: & Mean Robustness: & SD:      \\ \midrule
1      & 0.181474         & 0.1951   \\ \midrule
10     & 0.035735         & 0.007395 \\ \midrule
25     & 0.046072         & 0.005877 \\ \midrule
50     & 0.058456         & 0.006017 \\ \midrule
150    & 0.077806         & 0.005403 \\ \midrule
\textbf{300}    & \textbf{0.087175}         & \textbf{0.00396}  \\ \bottomrule
\end{tabular}
\end{center}
\end{table}

\newpage

\begin{table}[h]
\begin{center}
\caption{MNIST ResNet50 adversarial robustness}
\vspace{0.1in}
\label{tab:resnet-mnist}
\begin{tabular}{@{}lll@{}}
\toprule
Epoch: & Mean Robustness: & SD:      \\ \midrule
1      & 0.243206      & 0.064093 \\ \midrule
10     & 0.37482$\star\ddagger$        & 0.040255 \\ \midrule
25     & 0.413437$\star\ddagger$       & 0.037413 \\ \midrule
50     & 0.405367$\star\ddagger$      & 0.03901  \\ \midrule
100    & 0.486548$\star\ddagger$      & 0.015996 \\ \midrule
\textbf{125}    & \textbf{0.572381}$\star\ddagger$       & \textbf{0.03999}  \\ \bottomrule
\end{tabular}
\end{center}
\end{table}
\begin{table}[h]
\begin{center}
\caption{CIFAR-10 ResNet50 adversarial robustness}
\vspace{0.1in}
\label{tab:resnet-cifar}
\begin{tabular}{@{}lll@{}}
\toprule
Epoch: & Mean Robustness: & SD:      \\ \midrule
1      & 0.222479$\diamond$         & 0.098118 \\ \midrule
10     & 0.072652$\ddagger$         & 0.027665 \\ \midrule
25     & 0.06675$\ddagger$          & 0.010177 \\ \midrule
50     & 0.062354         & 0.007254 \\ \midrule
100    & 0.075903 $\ddagger$        & 0.00379  \\ \midrule
\textbf{125}    & \textbf{0.138936}$\diamond\ddagger$        & \textbf{0.038124} \\ \bottomrule
\end{tabular}
\end{center}
\end{table}

\begin{table}[h]
\begin{center}
\caption{Pretrained on CIFAR-10 trained on MNIST ResNet50 adversarial robustness}
\vspace{0.1in}
\label{tab:resnet-pre-cifar}
\begin{tabular}{@{}lll@{}}
\toprule
Epoch: & Mean Robustness: & SD:      \\ \midrule
1      & 0.392197$\bullet\diamond$         & 0.036562 \\ \midrule
10     & 0.422000$\bullet$         & 0.023966 \\ \midrule
25     & 0.414428$\bullet$         & 0.035434 \\ \midrule
50     & 0.413840$\bullet$         & 0.034093 \\ \midrule
100    & 0.491886$\bullet$         & 0.012084 \\ \midrule
\textbf{125 }   & \textbf{0.569465}$\bullet$         & \textbf{0.020692} \\ \bottomrule
\end{tabular}
\end{center}
\end{table}
\begin{table}[h]
\begin{center}
\caption{Pretrained on MNIST trained on CIFAR-10 ResNet50 adversarial robustness}
\vspace{0.1in}
\label{tab:resnet-pre-mnist}
\begin{tabular}{@{}lll@{}}
\toprule
Epoch: & Mean Robustness: & SD:      \\ \midrule
1      & 0.067825         & 0.008714 \\ \midrule
10     & 0.063501         & 0.0099   \\ \midrule
25     & 0.068075         & 0.012437 \\ \midrule
50     & 0.064227         & 0.010571 \\ \midrule
100    & 0.074815         & 0.004281 \\ \midrule
\textbf{125 }   & \textbf{0.114783 }        & \textbf{0.03279}  \\ \bottomrule
\end{tabular}
\end{center}
\end{table}

\begin{table}[h]
\begin{center}
\caption{Fusion digit task ResNet50 adversarial robustness}
\vspace{0.1in}
\label{tab:resnet-digit}
\begin{tabular}{@{}lll@{}}
\toprule
Epoch: & Mean Robustness: & SD:      \\ \midrule
1      & 0.252273         & 0.267929 \\ \midrule
10     & 0.180888$\dagger$         & 0.014562 \\ \midrule
25     & 0.190701$\dagger$         & 0.022835 \\ \midrule
50     & 0.202564$\dagger$        & 0.009819 \\ \midrule
100    & 0.242128$\dagger$         & 0.006506 \\ \midrule
\textbf{125}    & \textbf{0.296214}$\dagger$         & \textbf{0.018022} \\ \bottomrule
\end{tabular}
\end{center}
\end{table}
\begin{table}[h]
\begin{center}
\caption{Fusion object task ResNet50 adversarial robustness}
\vspace{0.1in}
\label{tab:resnet-object}
\begin{tabular}{@{}lll@{}}
\toprule
Epoch: & Mean Robustness: & SD:      \\ \midrule
1      & 0.243001         & 0.10571  \\ \midrule
10     & 0.053983         & 0.014983 \\ \midrule
25     & 0.045387         & 0.008189 \\ \midrule
50     & 0.065714         & 0.017707 \\ \midrule
100    & 0.070246         & 0.00476  \\ \midrule
\textbf{125}    & \textbf{0.084441 }        & \textbf{0.007446} \\ \bottomrule
\end{tabular}
\end{center}
\end{table}

\begin{table}[h]
\begin{center}
\caption{MNIST FullyConnectedNet adversarial robustness}
\vspace{0.1in}
\label{tab:fully-mnist}
\begin{tabular}{@{}lll@{}}
\toprule
Epoch: & Mean Robustness: & SD:     \\ \midrule
1      & 0.272342$\star\diamond\ddagger$        & 0.004514 \\ \midrule
10     & 0.280533$\star\diamond\ddagger$         & 0.002443 \\ \midrule
25     & 0.284781$\star\diamond\ddagger$        & 0.002756 \\ \midrule
50     & 0.29422$\star\diamond\ddagger$       & 0.002926 \\ \midrule
150    & 0.311588$\star\ddagger$        & 0.000868 \\ \midrule
\textbf{300 }   & \textbf{0.31289}6$\star\ddagger$        & \textbf{0.00083}  \\ \bottomrule
\end{tabular}
\end{center}
\end{table}
\begin{table}[h]
\begin{center}
\caption{CIFAR-10 FullyConnectedNet adversarial robustness}
\vspace{0.1in}
\label{tab:fully-cifar}
\begin{tabular}{@{}lll@{}}
\toprule
Epoch: & Mean Robustness: & SD:     \\ \midrule
1      & 0.086983$\ddagger$         & 0.003529 \\ \midrule
10     & 0.064134$\ddagger$         & 0.002695 \\ \midrule
25     & 0.054232$\ddagger$         & 0.00201  \\ \midrule
50     & 0.045977$\diamond\ddagger$         & 0.001153 \\ \midrule
150    & 0.037929$\diamond\ddagger$         & 0.000255 \\ \midrule
\textbf{300}    & \textbf{0.034145}$\diamond\ddagger$         & \textbf{0.000274} \\ \bottomrule
\end{tabular}
\end{center}
\end{table}

\begin{table}[h]
\begin{center}
\caption{Pretrained on CIFAR-10 trained on MNIST FullyConnectedNet adversarial robustness}
\vspace{0.1in}
\label{tab:fully-pre-cifar}
\begin{tabular}{@{}lll@{}}
\toprule
Epoch: & Mean Robustness: & SD:     \\ \midrule
1      & 0.153764$\bullet$         & 0.005824 \\ \midrule
10     & 0.194182$\bullet$         & 0.00334  \\ \midrule
25     & 0.235387$\bullet$         & 0.002786 \\ \midrule
50     & 0.281417$\bullet$         & 0.004191 \\ \midrule
150    & 0.320011$\bullet\diamond$         & 0.001317 \\ \midrule
\textbf{300}    & \textbf{0.321303}$\bullet\diamond$         & \textbf{0.001126} \\ \bottomrule
\end{tabular}
\end{center}
\end{table}
\begin{table}[h]
\begin{center}
\caption{Pretrained on MNIST trained on CIFAR-10 FullyConnectedNet adversarial robustness}
\vspace{0.1in}
\label{tab:fully-pre-mnist}
\begin{tabular}{@{}lll@{}}
\toprule
Epoch: & Mean Robustness: & SD:     \\ \midrule
1      & 0.104627$\diamond$         & 0.004495 \\ \midrule
10     & 0.069091$\diamond$         & 0.002008 \\ \midrule
25     & 0.053113         & 0.002535 \\ \midrule
50     & 0.044058         & 0.001362 \\ \midrule
150    & 0.036256        & 0.00035  \\ \midrule
\textbf{300}    & \textbf{0.031516}        & \textbf{0.000222} \\ \bottomrule
\end{tabular}
\end{center}
\end{table}

\begin{table}[h]
\begin{center}
\caption{Fusion digit task FullyConnectedNet adversarial robustness}
\vspace{0.1in}
\label{tab:fully-digit}
\begin{tabular}{@{}lll@{}}
\toprule
Epoch: & Mean Robustness: & SD:     \\ \midrule
1      & 0.113213$\dagger$         & 0.003298 \\ \midrule
10     & 0.129283$\dagger$         & 0.001951 \\ \midrule
25     & 0.139871$\dagger$         & 0.001997 \\ \midrule
50     & 0.145459$\dagger$         & 0.001856 \\ \midrule
150    & 0.15254$\dagger$          & 0.000776 \\ \midrule
\textbf{300}    & \textbf{0.15016}$\dagger$          & \textbf{0.000758} \\ \bottomrule
\end{tabular}
\end{center}
\end{table}
\begin{table}[h]
\begin{center}
\caption{Fusion object task FullyConnectedNet adversarial robustness}
\vspace{0.1in}
\label{tab:fully-object}
\begin{tabular}{@{}lll@{}}
\toprule
Epoch: & Mean Robustness: & SD:     \\ \midrule
1      & 0.052986         & 0.002406 \\ \midrule
10     & 0.036917         & 0.00154  \\ \midrule
25     & 0.03305          & 0.001239 \\ \midrule
50     & 0.030376         & 0.001014 \\ \midrule
150    & 0.026116         & 0.00027  \\ \midrule
\textbf{300}    & \textbf{0.022955}         & \textbf{0.000141} \\ \bottomrule
\end{tabular}
\end{center}
\end{table}

\end{document}